\def\year{2018}\relax
\documentclass[letterpaper]{article} 
\usepackage{aaai18}  
\usepackage{times}  
\usepackage{helvet}  
\usepackage{courier}  
\usepackage{url}  
\usepackage{graphicx}  

\usepackage{amsmath,amssymb}
\usepackage{color}
\usepackage{tabularx}
\usepackage[super]{nth}
\usepackage{etoolbox,siunitx}
\robustify\bfseries
\usepackage{booktabs}
\newcommand{\citet}[1]
{\citeauthor{#1}~\shortcite{#1}}
\newcommand{\citep}{\cite}

\begin{document}
%
\title{UnFlow: Unsupervised Learning of Optical Flow\\ with a Bidirectional Census Loss}
\author{Simon Meister, Junhwa Hur, Stefan Roth\\
Department of Computer Science\\ TU Darmstadt, Germany
}


\newcommand{\todo}[1]{\textbf{\textcolor{red}{#1}}}
\newcommand{\etal}{\textit{et al}.}
\newcommand{\ie}{\textit{i}.\textit{e}.}
\newcommand{\eg}{\textit{e}.\textit{g}.}
\newcommand{\cf}{\textit{cf}.}


\hyphenation{Conv-Net Conv-Nets}

\maketitle

\begin{abstract}
In the era of end-to-end deep learning, many advances in computer vision are driven by large amounts of labeled data.
In the optical flow setting, however, obtaining dense per-pixel ground truth for real scenes is difficult and thus such data is rare.
Therefore, recent end-to-end convolutional networks for optical flow rely on synthetic datasets for supervision, but the domain mismatch between training and test scenarios continues to be a challenge.
Inspired by classical energy-based optical flow methods, we design an unsupervised loss based on occlusion-aware bidirectional flow estimation and the robust census transform to circumvent the need for ground truth flow.
On the KITTI benchmarks, our unsupervised approach outperforms previous unsupervised deep networks by a large margin, and is even more accurate than similar supervised methods trained on synthetic datasets alone.
By optionally fine-tuning on the KITTI training data, our method achieves competitive optical flow accuracy on the KITTI 2012 and 2015 benchmarks, thus in addition enabling generic pre-training of supervised networks for datasets with limited amounts of ground truth.
\end{abstract}

\section{Introduction}
\label{sec:introduction}

Estimating dense optical flow is one of the longstanding problems in computer vision, with a variety of applications.
Though numerous approaches have been proposed over the past decades, \eg~\cite{Black:1991:RDM,Bruhn:2005:LKM,Brox:2011:LDO,Revaud:2015:EEP},
realistic benchmarks such as MPI Sintel \cite{Butler:2012:NOS} or KITTI \cite{Geiger:2012:AWR,Menze:2015:OSF} continue to challenge traditional energy minimization approaches.
Motivated by the recent successes of end-to-end deep learning,
convolutional neural networks (CNNs) have been suggested for addressing these challenges,
and recently started to outperform many traditional flow methods in public benchmarks \cite{Dosovitskiy:2015:FLO,Gadot:2016:PBB,Gueney:2016:DDF,Ilg:2017:FEO}.

Still, most CNN-based methods rely on the availability of a large amount of data with ground truth optical flow for supervised learning.
Due to the difficulty of obtaining ground truth in real scenes, such networks are trained on synthetically generated images, for which dense ground truth is easier to obtain in large amounts \cite{Dosovitskiy:2015:FLO,Mayer:2016:LDT}.
However, because of the intrinsic differences between synthetic and real imagery and the limited variability of synthetic datasets, the generalization to real scenes remains challenging.
We posit that for realizing the potential of deep learning in optical flow,
addressing this issue will be crucial.

In order to cope with the lack of labeled real-world training data, recent work \cite{Ahmadi:2016:UCN,Yu:2016:BBU,Ren:2017:UDL} has proposed optical flow networks based on unsupervised learning.
They sidestep the need for ground truth motion as only image sequences are required for training.
Though unsupervised learning seems to be a promising direction for breaking the dependency on synthetic data, one may argue that these approaches fall short of the expectations so far, as they neither match nor surpass the accuracy of supervised methods despite the domain mismatch between training and test time.

\begin{figure*}
\centering
\includegraphics[width=\textwidth]{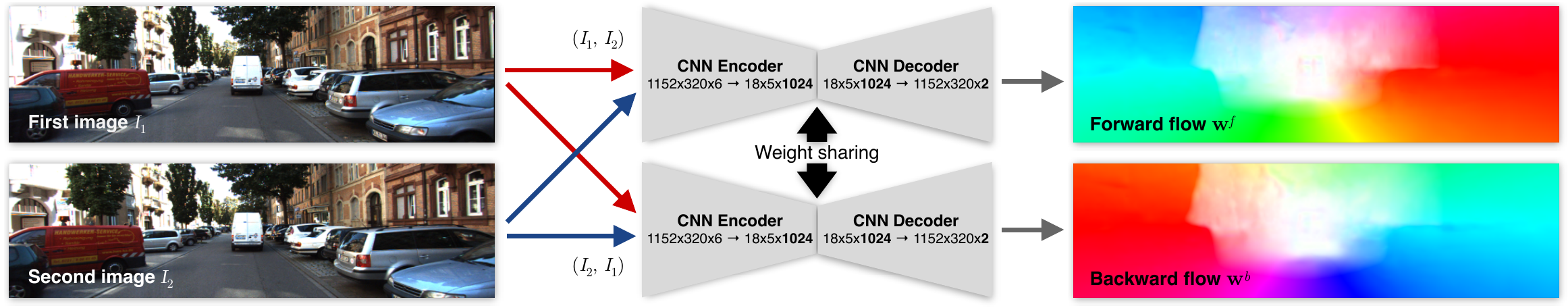}
\caption{
Bidirectional training of FlowNet \cite{Dosovitskiy:2015:FLO} with our comprehensive unsupervised loss.
}
\label{figure:teaser}
\end{figure*}

In this paper, we introduce an end-to-end unsupervised approach that demonstrates the effectiveness of unsupervised learning for optical flow.
Building on recent optical flow CNNs \cite{Dosovitskiy:2015:FLO,Ilg:2017:FEO},
we replace the supervision from synthetic data by an unsupervised photometric reconstruction loss similar to \cite{Yu:2016:BBU}.
We compute bidirectional optical flow (\ie, both in forward and backward direction, see Fig.~\ref{figure:teaser}) by performing a second pass with the two input images exchanged.
Then, we design a loss function leveraging bidirectional flow to explicitly reason about occlusion \cite{Hur:2017:MFE} and make use of the census transform to increase robustness on real images.
Through a comprehensive ablation study, we validate the design choices behind our unsupervised loss.

On the challenging KITTI benchmarks \cite{Geiger:2012:AWR,Menze:2015:OSF},
our unsupervised model outperforms previous unsupervised deep networks by a very large margin.
Perhaps more surprisingly, it also surpasses architecturally similar supervised approaches trained exclusively on synthetic data.
Similar observations hold, for example, on the Middlebury dataset \cite{Baker:2011:DBE}.
After unsupervised training on a large amount of realistic domain data, we can optionally make use of sparse ground truth (if available) to refine our estimates in areas that are inherently difficult to penalize in an unsupervised
way, such as at motion boundaries.
Our fine-tuned model achieves leading accuracy on KITTI, thus demonstrating that our unsupervised learning approach also enables pre-training for supervised methods on domains with limited amounts of ground truth data.
We thus present a step toward eliminating the need for careful engineering of synthetic datasets, as our networks can be trained on other domains with the expectation of achieving competitive accuracy.

\section{Related Work}
\label{sec:relatedwork}

\subsubsection{End-to-end supervised learning.}
End-to-end supervised learning of convolutional networks for optical flow was first introduced with FlowNet \cite{Dosovitskiy:2015:FLO}.
The network takes two consecutive input images and outputs a dense optical flow map using an encoder-decoder architecture.
For supervised training, a large synthetic dataset was generated from static background images and renderings of animated 3D chairs.
However, most likely due to the limited realism of the training dataset, FlowNet \cite{Dosovitskiy:2015:FLO} performs worse on the realistic KITTI 2012 benchmark \cite{Geiger:2012:AWR} when compared to synthetic benchmarks like MPI Sintel \cite{Butler:2012:NOS}.

A follow up work \cite{Ilg:2017:FEO} introduced the more accurate, but also slower and more complex FlowNet2 family of supervised networks.
They improve upon the original architecture by stacking multiple FlowNet networks for iterative refinement.
In addition to the original synthetic data, a more complex synthetic dataset \cite{Mayer:2016:LDT} featuring a larger variety of objects and motions is used for training.
For the KITTI benchmark, FlowNet2 is fine-tuned on the sparse ground truth from the KITTI 2012 \cite{Geiger:2012:AWR}
and KITTI 2015 \cite{Menze:2015:OSF} training sets and achieves state-of-the-art accuracy.

A number of other deep architectures based on supervised learning have been proposed, leveraging various paradigms, including patch matching \cite{Gadot:2016:PBB}, discrete optimization \cite{Gueney:2016:DDF}, and coarse-to-fine estimation \cite{Ranjan:2017:OFE}.
While we focus on the FlowNet architecture and its variants here, we note that our unsupervised loss can, in principle, be combined with other network architectures that directly predict the flow, \eg~ \cite{Ranjan:2017:OFE}.

\subsubsection{End-to-end unsupervised learning.}
Recently, \citet{Yu:2016:BBU} and \citet{Ren:2017:UDL} suggested an unsupervised method based on FlowNet for sidestepping the limitations of synthetic datasets.
As first proposed in earlier work with a different network architecture \cite{Ahmadi:2016:UCN}, they replace the original supervised loss by a proxy loss based on the classical brightness constancy and smoothness assumptions.
The unsupervised network is trained on unlabeled image pairs from videos provided by the raw KITTI dataset \cite{Geiger:2013:VMR}.
On the KITTI 2012 benchmark \cite{Geiger:2012:AWR}, the unsupervised method fails to outperform the original FlowNet, probably due to the proxy loss being too simplistic.
Subsequent work attempted to overcome these problems by combining the unsupervised proxy loss with proxy ground truth generated by a classical optical flow algorithm \cite{Zhu:2017:GOF}.
However, the gain on previous purely unsupervised approaches is minor, falling short of the original supervised FlowNetS.
Other subsequent work \cite{Zhu:2017:DDF} attempted to improve unsupervised CNNs by replacing the underlying FlowNet architecture with a different network.
Again, the method shows only little improvement over \cite{Yu:2016:BBU,Ren:2017:UDL} and is still outperformed by the supervised FlowNetS.

As prior work does not come close to the accuracy of supervised methods, it remains unclear if unsupervised learning can overcome some of the limitations of supervised methods in realistic scenarios.
In this paper, we therefore investigate possible ways for improving the accuracy of unsupervised approaches
and aim to uncover whether they are a viable alternative or addition to supervised learning.

\newcommand{\Vwf}{\mathbf{w}^f}
\newcommand{\Vwb}{\mathbf{w}^b}
\newcommand{\Vx}{\mathbf{x}}
\newcommand{\Vr}{\mathbf{r}}
\newcommand{\Vs}{\mathbf{s}}

\section{Unsupervised Learning of Optical Flow}
\label{sec:approach}

We build on the previous FlowNetS-based UnsupFlownet \cite{Yu:2016:BBU} and extend it in three important ways.
First, we design a symmetric, occlusion-aware loss based on bidirectional (\ie, forward and backward) optical flow estimates.
Second, we train FlowNetC with our comprehensive unsupervised loss to estimate bidirectional flow.
Third, we make use of iterative refinement by stacking multiple FlowNet networks \cite{Ilg:2017:FEO}.
Optionally, we can also use a supervised loss for fine-tuning our networks on sparse ground truth data after unsupervised training.

\subsection{Unsupervised loss}

\begin{figure*}
\centering
\includegraphics[width=\textwidth]{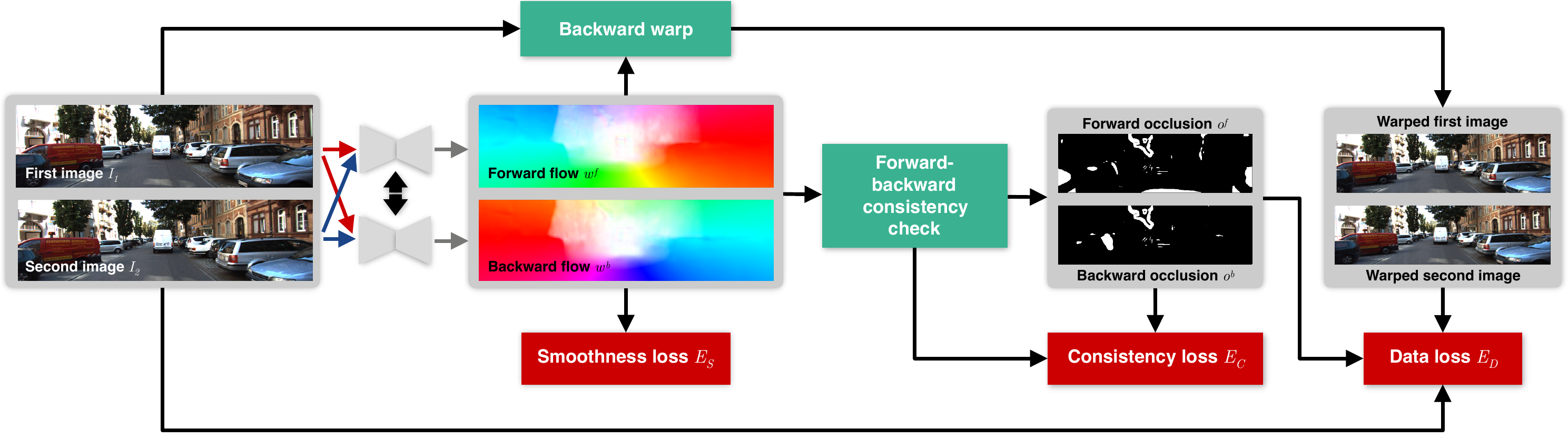}
\caption{
Schematic of our unsupervised loss.
The data loss compares flow-warped images to the respective original images and penalizes their difference.
We use forward-backward consistency based on warping the flow fields for estimating occlusion maps, which mask
the differences in the data loss.
Both flow fields are regularized assuming $2^{\text{nd}}$-order smoothness.
}
\label{figure:unsupervised_loss}
\end{figure*}

Let $I_1,I_2 : P \to \mathbb{R}^3$ be two temporally consecutive frames.
Our goal is to estimate the optical flow $\Vwf = (u^f, v^f)^T$ from $I_1$ to $I_2$.
As our occlusion detection additionally requires the inverse (backward) optical flow $\Vwb = (u^b, v^b)^T$,
we jointly estimate bidirectional flow by making all of our loss terms symmetrical
(\ie, computing them for both flow directions).
Bidirectional flow estimation taking into account occlusions has been proposed in the context of classical energy-based optical flow \cite{Hur:2017:MFE}.
Here, we extend this idea from a superpixel-based setting to general flow fields and apply it as a loss function in unsupervised learning.
Note that for brevity, some textual explanations discuss the forward direction only,
however the same reasoning will apply to the backward direction as well.

Our unsupervised loss is based on the observation that a pixel in the first frame
should be similar to the pixel in the second frame to which it is mapped by the flow \cite{Fleet:2006:OFE},
which we encourage with our data loss.
This observation does not hold for pixels that become occluded, however,
as the corresponding pixels in the second frame are not visible.
Thus, we mask occluded pixels from the data loss to avoid
learning incorrect deformations that fill in the occluded pixels \cite{Xiao:2006:BFO}.
Our occlusion detection is based on the forward-backward consistency assumption \cite{Sundaram:2010:DPT}.
That is, for non-occluded pixels, the forward flow should be the inverse of the backward flow at the corresponding pixel in the second frame.
We mark pixels as becoming occluded whenever the mismatch between these two flows is too large.
Thus, for occlusion in the forward direction, we define the occlusion flag
$o_{\Vx}^f$ to be $1$ whenever the constraint
\begin{multline}
  \Big|\Vwf(\Vx) + \Vwb\big(\Vx + \Vwf(\Vx)\big)\Big|^2 \\
  < \alpha_1\Bigg(\Big|\Vwf(\Vx)\Big|^2 + \Big|\Vwb\big(\Vx + \Vwf(\Vx)\big)\Big|^2\Bigg) + \alpha_2
\label{eq:occ_detect}
\end{multline}
is violated, and $0$ otherwise.
For the backward direction, we define $o_{\Vx}^b$ in the same way with $\Vwf$ and $\Vwb$ exchanged.
We set $\alpha_1=0.01$, $\alpha_2=0.5$ in all of our experiments.
As a baseline, we will also explore a loss variant with occlusion masking disabled.
In that case, we simply let $o_{\Vx}^f = o_{\Vx}^b = 0$ for all $\Vx \in P$.

Our occlusion-aware data loss is now defined as
\begin{multline}
E_D(\Vwf, \Vwb, o^f, o^b) = \\
  \sum_{\Vx \in P}
  (1 - o_{\Vx}^f) \cdot
  \rho \Big( f_D \big(I_1(\Vx), I_2(\Vx + \Vwf(\Vx)) \big) \Big) \\
  \;\;\,+ (1 - o_{\Vx}^b) \cdot
  \rho \Big( f_D \big(I_2(\Vx), I_1(\Vx + \Vwb(\Vx)) \big) \Big) \\
  + {o_{\Vx}^f}\lambda_p  + {o_{\Vx}^b}\lambda_p,
\label{eq:data_term}
\end{multline}
where $f_D(I_1(\Vx), I_2(\Vx'))$ measures the photometric difference between two putatively corresponding pixels $\Vx$ and $\Vx'$, and $\rho(x) = (x^2 + \epsilon^2)^\gamma$ is the robust generalized Charbonnier penalty function \cite{Sun:2014:QAC}, with $\gamma = 0.45$.
We add a constant penalty $\lambda_p$ for all occluded pixels to avoid the trivial solution where all pixels become occluded, and penalize the photometric difference for all non-occluded pixels.
In previous work \cite{Yu:2016:BBU}, the brightness constancy constraint $f_D(I_1(\Vx), I_2(\Vx')) = I_1(\Vx) - I_2(\Vx')$ was used for measuring the photometric difference.
As the brightness constancy is not invariant to illumination changes common in realistic situations \cite{Vogel:2013:EDC}, we instead use the ternary census transform \cite{Zabih:1994:NPL,Stein:2004:ECO}.
The census transform can compensate for additive and multiplicative illumination changes as well as changes to gamma \cite{Hafner:2013:WCT},
thus providing us with a more reliable constancy assumption for realistic imagery.

Unlike \citet{Yu:2016:BBU}, we use a second-order smoothness constraint \cite{Trobin:2008:USO,Zhang:2014:ARA} on the flow field to encourage collinearity of neighboring flows
and thus achieve more effective regularization:
\begin{multline}
E_S(\Vwf, \Vwb) = \\
  \;\sum_{\Vx \in P} \sum_{(\Vs, \Vr) \in N(\Vx)}
  \rho\Big(\Vwf(\Vs) - 2 \Vwf(\Vx) + \Vwf(\Vr)\Big) \\
 + \rho\Big(\Vwb(\Vs) - 2 \Vwb(\Vx) + \Vwb(\Vr)\Big),
\label{eq:smoothness}
\end{multline}
where $N(\Vx)$ consists of the horizontal, vertical, and both diagonal neighborhoods around $\Vx$ (4 in total).
For vectorial arguments, we assume that $\rho(\cdot)$ computes the average over the original generalized Charbonnier penalties of each component.
Note that for occluded pixel positions, this term is the only one active besides the occlusion penalty.

For non-occluded pixels, we add a forward-backward consistency penalty
\begin{multline}
E_C(\Vwf, \Vwb, o^f, o^b) = \\
  \;\;\sum_{\Vx \in P} (1 - o_{\Vx}^f)  \cdot \rho\Big(\Vwf(\Vx) + \Vwb\big(\Vx + \Vwf(\Vx)\big)\Big)\\
  + (1 - o_{\Vx}^b)  \cdot \rho\Big(\Vwb(\Vx) + \Vwf\big(\Vx + \Vwb(\Vx)\big)\Big).
\label{eq:consistency}
\end{multline}

Then, our final loss is a weighted sum of the individual loss terms
\begin{multline}
E(\Vwf, \Vwb, o^f, o^b) = E_D(\Vwf, \Vwb, o^f, o^b) \\
  + \lambda_S E_S(\Vwf, \Vwb) + \lambda_C E_C(\Vwf, \Vwb, o^f, o^b).
\label{eq:1}
\end{multline}
The interplay of all losses is illustrated in Fig.~\ref{figure:unsupervised_loss}.

\subsubsection{Backward warping.}
To compute our losses in a subdifferentiable way for use with backpropagation,
we employ bilinear sampling at flow-displaced positions (\ie, backward warping).
For example, to compare $I_1(\Vx)$ and $I_2(\Vx + \Vwf(\Vx))$, we backward-warp $I_2$ using $\Vwf$
as described by \citet{Yu:2016:BBU}
and then compare the backward-warped second image to the first image.
We use the bilinear sampling scheme of \citet{Jadeberg:2015:STN} to whom we refer for details.

\subsubsection{Loss differentiation and implementation.}
We implement all losses as well as the warping scheme with primitive TensorFlow functions \cite{Abadi:2015:TLM}
and use automatic differentiation for backpropagation.
Source code for training and evaluating our models is publicly available.

\subsection{Network architecture and computation}

\subsubsection{UnFlow-C.}
Our basic CNN, termed \emph{UnFlow-C}, is based on FlowNetC \cite{Dosovitskiy:2015:FLO},
which processes two consecutive images in two separate input streams, explicitly correlates them,
and compresses the result with a CNN encoder down to a sixth of the original resolution.
In a decoder part (\emph{refinement network}), the compressed representation is convolutionally upsampled four times as in FlowNetC, and dense flow
is predicted after each upsampling.
The last flow estimate is then bilinearly upsampled to the original resolution.
To compute bidirectional optical flow, we first apply FlowNetC to the RGB images $(I_1, I_2)$
to obtain the forward flow $(u^f, v^f)$ and apply the same computation to $(I_2, I_1)$ to obtain the backward flow $(u^b, v^b)$.
We share weights between both directions to train a universal network for optical flow in either direction (see Fig.~\ref{figure:teaser}).

\subsubsection{Stacking.}
Inspired by FlowNet2 \cite{Ilg:2017:FEO}, we iteratively refine the estimate of
\emph{UnFlow-C} by passing it into a separate FlowNetS with independent weights
and term the two-network stack \emph{UnFlow-CS}.
Again, we perform one pass for each flow direction and share all weights between the two passes.
In addition to the original images, we input the initial flow estimate,
the backward-warped second image, and the brightness error between the warped image and the first image into the iterated network.
In the same way, we concatenate an additional FlowNetS after UnFlow-CS
to refine its estimate and term the three-network stack \emph{UnFlow-CSS}.

\subsubsection{Computing the unsupervised loss.}
Similar to the supervised FlowNet,
we compute losses for all intermediate predictions
(\cf~ Fig.~3 in \cite{Dosovitskiy:2015:FLO} or Fig.~2 in \cite{Yu:2016:BBU})
from the refinement network to guide the learning process
at multiple resolutions and then combine them by taking a weighted average.
Our total loss is

\begin{equation}
E_{\mathrm{unsup}} = \sum_i \lambda^i_l E_i,
\label{eq:final_loss}
\end{equation}
where $E_i$ is the loss from Eq.~\eqref{eq:1}, evaluated at layer $i$.
Table \ref{table:losslayers} details the settings we use for each individual layer.

{
\begin{table}[t]
\centering
\begin{tabular}{@{}ccS[table-format=2.2]c@{}}
\toprule
{$i^\text{th}$ layer} & {Scale} & {Weight $\lambda^i_l$} & {Patch size} \\ \midrule
  0 & $2^{-6}$ & 1.1 & $3\times3$ \\
  1 & $2^{-5}$ & 3.4 & $3\times3$ \\
  2 & $2^{-4}$ & 3.9 & $5\times5$ \\
  3 & $2^{-3}$ & 4.35 & $5\times5$ \\
  4 & $2^{-2}$ & 12.7 & $7\times7$ \\
  \bottomrule
\end{tabular}
\caption{
Loss settings to penalize flow fields predicted at different stages of the
refinement network (for the final, highest resolution estimate, $i = 4$).
The scale of the respective estimate is given as a fraction of the input image resolution
and the loss weight decreases with the resolution of the estimate.
We use a smaller patch size for the census transform as resolution decreases.
}
\label{table:losslayers}
\end{table}
}

\subsubsection{Supervised loss for fine-tuning.}
For supervised fine-tuning on the KITTI training sets
(only relevant for network variants with suffix \emph{-ft}),
we compute the network loss by comparing the bilinearly upsampled
final flow estimate to the ground truth flow at all pixels for which ground truth is available:
\begin{equation}
E_{\mathrm{sup}}(\Vwf) = \sum_{\Vx \in P} v_{\Vx}^f \rho\big(\Vwf(\Vx) - \Vwf_{gt}(\Vx)\big),
\end{equation}
where $v_{\Vx}^f = 1$ if there is valid ground truth at pixel $\Vx$ and $v_{\Vx}^f = 0$ otherwise.
As we want to avoid further increasing the sparsity of the ground truth,
we do not downsample the ground truth flow and compute the loss for the final prediction only.
Note that during fine-tuning, we only compute the first pass of the network for
the forward flow direction, as there is ground truth for this direction only.

\subsection{Discussion}
As opposed to supervised approaches, which are often limited by the amount of available ground truth data, unsupervised techniques such as ours are limited by the loss function.
That is, it should be noted that the amount of information that can be gained from the available video data is limited by how faithfully the problem is modeled by the loss.
This is the key challenge that we are trying to address here.

Another limitation of our method is that we need to perform a parameter search for the weighting between the loss terms (\eg, the influence of the smoothness loss) for best results, which increases the total computation time for training a model on a new domain for the first time.
This limitation is shared by previous works employing an unsupervised proxy loss \cite{Ahmadi:2016:UCN,Yu:2016:BBU,Ren:2017:UDL}.

Compared to standard energy-based optical flow methods, our unsupervised networks avoid expensive optimization at test time.
Moreover, stochastic minimization of the loss over a whole (training) dataset, as done here, can avoid some of the pitfalls of optimizing a complex energy on individual inputs, as done in classical approaches.

\begin{figure*}[t]
  \centering
  \includegraphics[width=\textwidth]{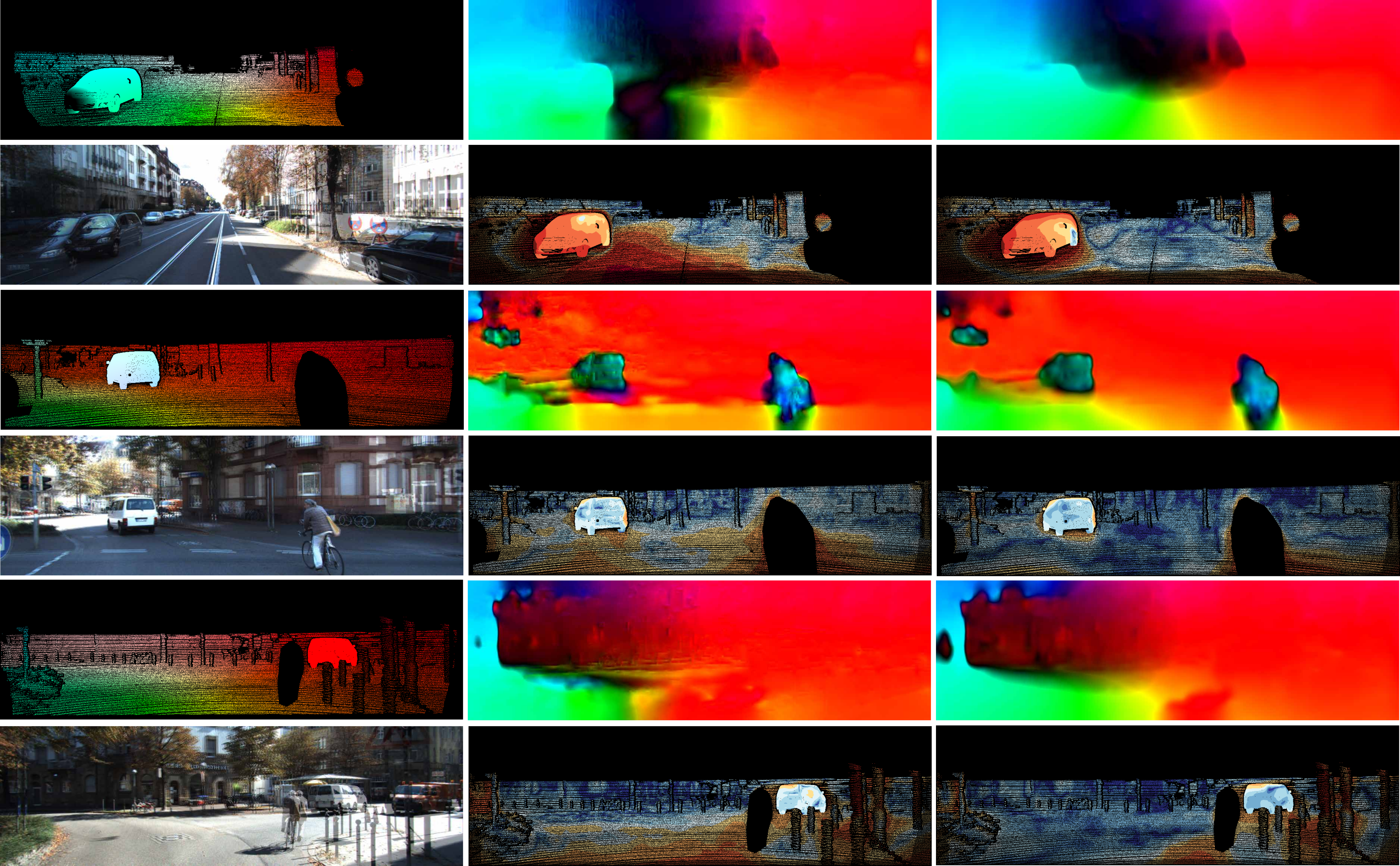}
\caption{
Visual comparison of our fully unsupervised \mbox{UnFlow-C} on the KITTI 2015 training set.
We compare a baseline unsupervised loss akin to \cite{Yu:2016:BBU} (middle) to our best unsupervised loss (right).
For each example, we show the ground truth and estimated flow (upper row), as well as input image overlay and flow error (lower row).
The KITTI 2015 error map depicts correct estimates ($\leq 3$ px or $\leq 5\%$ error) in blue and wrong estimates in red tones.
}
\label{figure:losscomparison}
\end{figure*}

\section{Datasets for Training}
\label{sec:Datasets}

{
\begin{table*}[t]
\centering
\resizebox{\textwidth}{!}{
\begin{tabular}{@{}cccS[table-format=1.1]S[table-format=2.1]S[table-format=1.1]S[table-format=1.2]S[table-format=1.2]S[table-format=2.2,table-space-text-post=\%]S[table-format=2.2]S[table-format=1.2]S[table-format=2.2,table-space-text-post=\%]@{}}
\toprule
\multicolumn{6}{@{}c}{Loss parameters} & \multicolumn{3}{c}{KITTI 2012} & \multicolumn{3}{c@{}}{KITTI 2015} \\
\cmidrule(lr){1-6}\cmidrule(lr){7-9}\cmidrule(lr){10-12}
$f_D$ & occ. masking & $E_S$ & {$\lambda_S$} & {$\lambda_p$} & {$\lambda_C$}
  & {AEE (All)} & {AEE (NOC)} & {Fl-all} & {AEE (All)} & {AEE (NOC)} & {Fl-all} \\\midrule
  $\textsuperscript{\textdagger}$
  brightness & $\times$ & \nth{1} & 3.0 & 0.0 & 0.0 &     8.0  & 4.1  & {--}       & 15.5  & 8.6  & {--} \\
  \midrule
  brightness & $\times$ & \nth{1} & 3.0 & 0.0 & 0.0 &     7.20 & 3.42 & 31.93\si{\percent} & 14.34 & 7.35 & 41.43\si{\percent} \\
  brightness & $\times$ & \nth{2} & 3.0 & 0.0 & 0.0 &     7.11 & 3.03 & 28.10\si{\percent} & 14.17 & 6.67 & 38.64\si{\percent} \\
  census & $\times$ & \nth{1} & 3.0 & 0.0 & 0.0 &         4.66 & 1.83 & 20.85\si{\percent} & 10.24 & 4.64 & 31.33\si{\percent} \\
  census & $\times$ & \nth{2} & 3.0 & 0.0 & 0.0 &         4.40 & 1.63 & 17.22\si{\percent} & 9.49 & \bfseries 4.10 & 29.20\si{\percent} \\
  census & \checkmark & \nth{2} & 3.0 & 12.4 & 0.0 &      4.32 & 1.62 & 17.14\si{\percent} & 9.17 & 4.40 & 29.21\si{\percent} \\
  census & \checkmark & \nth{2} & 3.0 & 12.4 & 0.2 &      \bfseries 3.78 & \bfseries 1.58 & \bfseries 16.44\si{\percent} & \bfseries 8.80 & 4.29 & \bfseries 28.95\si{\percent} \\
\bottomrule
\end{tabular}
}
\caption {
Comparison of different loss terms on the training sets of the KITTI benchmarks.
AEE: Average Endpoint Error; Fl-all: Ratio of pixels where flow estimate is
wrong by both $\geq 3$ pixels and $\geq 5\%$.
We report the AEE over all pixels (All) and over non-occluded pixels only (NOC).
The best result for each metric is printed in bold.
$\textsuperscript{\textdagger}\mathrm{For}$ the first row only, we train following the schedule of \citet{Yu:2016:BBU},
with pre-training done on FlyingChairs instead of SYNTHIA.
}
\label{table:losscomparison}
\end{table*}
}

\subsubsection{SYNTHIA.}
\citet{Ilg:2017:FEO} showed that pre-training FlowNet on the synthetic FlyingChairs dataset before training on more complex and realistic datasets consistently improves the accuracy of their final networks.
They conjecture that this pre-training
can help the networks to learn general concepts of optical flow before learning to handle
complex lighting and motion.
Motivated by this, we pre-train our unsupervised models on the synthetic SYNTHIA dataset \cite{Ros:2016:TSD},
for which no optical flow ground truth is available.
The SYNTHIA dataset consists of multiple views recorded from a vehicle driving through a virtual environment.
We use left images from the front, back, left, and right views of all winter and summer driving sequences,
which amount to about 37K image pairs.
Note that we can, of course, also pre-train on Flying Chairs (with comparable results after the full training procedure),
but we specifically avoid this to show that we do not depend on this dataset.

\subsubsection{KITTI.}
The KITTI dataset \cite{Geiger:2013:VMR} consists of real road scenes captured by a car-mounted stereo camera rig.
A laser scanner provides accurate yet sparse optical flow ground truth for a small number of images,
which make up the KITTI 2012 \cite{Geiger:2012:AWR} and KITTI 2015 \cite{Menze:2015:OSF} flow benchmarks.
In addition, a large dataset of raw $1392\times512$ image sequences is provided without ground truth.
For unsupervised training, we use pairs of contiguous images from the city, residential, and road categories of the raw dataset.
To avoid training on images on which we evaluate, we remove all images included in the KITTI 2012 and 2015 train and test sets
from the raw dataset, including their temporal neighbors (within $\pm 10$ frames).
This final dataset consists of about 72K image pairs.
For \emph{optional} supervised fine-tuning, we use the 394 image pairs with ground truth from the KITTI 2012 and KITTI 2015
training sets.
We evaluate our networks on the 195 testing and 194 training pairs of KITTI 2012 and
the 200 testing and 200 training pairs of KITTI 2015.

\subsubsection{Cityscapes.}
The Cityscapes dataset \cite{Cordts:2016:CDS} contains real driving sequences annotated for semantic segmentation and instance segmentation, without optical flow ground truth.
In addition to our main experiments on KITTI, we also train a model on Cityscapes.
We use all consecutive frames from the train/val/test sequences.
As the KITTI benchmarks are at 10Hz, while Cityscapes is recorded at 17Hz, we also include all pairs where
each second frame is skipped to learn larger displacements at 8.5Hz.
The final dataset consists of about 230K image pairs.

\section{Experiments}
\label{sec:Experiments}

\subsection{Training and evaluation details}
Our networks are first trained on SYNTHIA and then on KITTI raw or Cityscapes in the proposed unsupervised way.
We then \emph{optionally} apply supervised fine-tuning to our best stacked networks trained on KITTI,
for which we use ground truth from the KITTI 2012 and 2015 training sets.
As optimizer, we use Adam \cite{Kingma:2015:AMS} with $\beta_1 = 0.9$ and $\beta_2 = 0.999$.

\subsubsection{Unsupervised SYNTHIA pre-training.}
We train for 300K iterations with a mini-batch size of 4 image pairs from the SYNTHIA data.
We keep the initial learning rate of 1.0e--4 fixed for the first 100K iterations and then divide it by two
after every 100K iterations.
When training stacked networks, we pre-train each network while keeping the weights of any previous networks fixed.

\subsubsection{Unsupervised KITTI training.}
We train for 500K iterations with a mini-batch size of 4 image pairs from the raw KITTI data.
We keep the initial learning rate of 1.0e--5 fixed for the first 100K iterations and then divide it by two
after every 100K iterations.
For stacking, the same method as for SYNTHIA is applied.

\subsubsection{Unsupervised Cityscapes training.}
The procedure is the same as for KITTI, expect that we only train a single network without stacking.

\subsubsection{Supervised KITTI fine-tuning} (\emph{-ft})\textbf{.}
We fine-tune with a mini-batch size of 4 image pairs from the KITTI training sets and an initial learning rate of 0.5e--5,
which we reduce to 0.25e--5 and 0.1e--5 after 45K and 65K iterations, respectively.
For validation, we set aside 20\% of the shuffled training pairs and fine-tune
until the validation error increases, which generally occurs after about 70K iterations.
Fine-tuning of a stack is done end-to-end.

\subsubsection{Pre-processing and data augmentations during training.}
First, we shuffle the list of image pairs, and then
randomly crop KITTI (both, raw and training) images to $1152\times320$, SYNTHIA images to $768\times512$,
and Cityscapes images to $1024\times512$.
Following \citet{Yu:2016:BBU}, we apply random additive Gaussian noise ($0\!<\!\sigma\!\leq\! 0.04$),
random additive brightness changes, random multiplicative color changes ($0.9\!\leq\!\text{multiplier}\!\leq\! 1.1$),
as well as contrast (from $[-0.3,0.3]$) and gamma changes (from $[0.7,1.5]$)
to both frames independently.
For unsupervised training only, we first apply the same random horizontal flipping and scaling ($0.9\!\leq\!\text{factor}\!\leq\! 1.1$) to both frames, and finally a relative random scaling of the second frame ($0.9\!\leq\!\text{factor}\!\leq\! 1.1$).
The brightness changes are sampled from a Gaussian with $\sigma = 0.02$, while all other
random values are uniformly sampled from the given ranges.

{
\begin{table*}[t]
\centering
\resizebox{\textwidth}{!}{
\begin{tabular}{@{}lS[table-format=3.3]S[table-format=2.1]S[table-format=2.3]S[table-format=1.1]S[table-format=3.3]S[table-format=3.2,table-space-text-post={\%~~)}]S[table-format=2.2,table-space-text-post=\%]S[table-format=1.2]S[table-format=1.2]S[table-format=3.3]S[table-format=2.2]@{}}
\toprule
& \multicolumn{4}{c}{KITTI 2012} & \multicolumn{3}{c}{KITTI 2015} & \multicolumn{2}{c}{Middlebury} & \multicolumn{2}{c}{Sintel Final} \\
\cmidrule(lr){2-5}\cmidrule(lr){6-8}\cmidrule(lr){9-10}\cmidrule(lr){11-12}
Method & \multicolumn{2}{c}{AEE (All)} & \multicolumn{2}{c}{AEE (NOC)} & {AEE (All)} & \multicolumn{2}{c}{Fl-all} & \multicolumn{2}{c}{AEE} & \multicolumn{2}{c}{AEE} \\
\cmidrule(lr){2-3}\cmidrule(lr){4-5}\cmidrule(lr){6-6}\cmidrule(lr){7-8}\cmidrule(lr){9-10}\cmidrule(lr){11-12}
& {\emph{train}} & {\emph{test}} & {\emph{train}} & {\emph{test}} & {\emph{train}} & {\emph{train}} & {\emph{test}} & {\emph{train}} & {\emph{test}} & {\emph{train}} & {\emph{test}} \\ \midrule
	DDF {\footnotesize\cite{Gueney:2016:DDF}}	& {--} & 3.4 & {--} & 1.4 & {--} & {--} & 21.17\si{\percent} & {--} & {--} & {--} & 5.73 \\
	PatchBatch {\footnotesize\cite{Gadot:2016:PBB}}	& {--} & 3.3 & {--} & 1.3 & {--} & {--} & 21.07\si{\percent} & {--} & {--} & {--} & 5.36 \\
	FlowFieldCNN {\footnotesize\cite{Bailer:2017:CPM}}	& {--} & 3.0 & {--} & 1.2 & {--} & {--} & 18.68\si{\percent} & {--} & {--} & {--} & {--} \\
	ImpPB+SPCI {\footnotesize\cite{Schuster:2017:OFR}}	& {--} & 2.9 & {--} & 1.1 & {--} & {--} & 17.78\si{\percent} & {--} & {--} & {--} & {--} \\
	SDF {\footnotesize\cite{Bai:2016:ESI}}	& {--} & 2.3 & {--} & 1.0 & {--} & {--} & 11.01\si{\percent} & {--} & {--} & {--} & {--} \\ \hline
	FlowNetS+ft {\footnotesize\cite{Dosovitskiy:2015:FLO}}	 & 7.5 & 9.1 & 5.3 & 5.0 & {--} & {--} & {--} & 0.98 & {--} & (4.44) & 7.76 \\
	UnsupFlownet {\footnotesize\cite{Yu:2016:BBU}}			     & 11.3 & 9.9 & 4.3 & 4.6 & {--} & {--} & {--} & {--} & {--} & {--} & {--} \\
	DSTFlow(KITTI) {\footnotesize\cite{Ren:2017:UDL}}			 & 10.43 & 12.4 & 3.29 & 4.0 & 16.79 & 36\si{\percent} & 39\si{\percent} & {--} & {--} & 7.95 & 11.80 \\
	FlowNet2-C {\footnotesize\cite{Ilg:2017:FEO}}		       & {--} & {--} & {--} & {--} & 11.36 & {--} & {--} & {--} & {--} & {--} & {--} \\
	FlowNet2-CSS {\footnotesize\cite{Ilg:2017:FEO}}		     & 3.55 & {--} & {--} & {--} & 8.94 & 29.77\si{\percent}{\textsuperscript{\textdagger}} & {--} & 0.44 & {--} & 3.23 & {--} \\
	FlowNet2-ft-kitti \cite{Ilg:2017:FEO}	   & (1.28) & 1.8 & {--} & 1.0 & (2.30) & (8.61\si{\percent}{)\textsuperscript{\textdagger}} & 10.41\si{\percent} & 0.56 & {--} & 4.66 & {--} \\ \hline
	\textbf{\mbox{UnFlow-C-Cityscapes} (ours)}	& 5.08 & {--} & 2.12 & {--} & 10.78 & 33.89\si{\percent} & {--} & 0.85 & {--} & 8.23 & {--} \\
	\textbf{\mbox{UnFlow-C} (ours)}				& 3.78 & {--} & 1.58 & {--} & 8.80 & 28.94\si{\percent} & {--} & 0.88 & {--} & 8.64 & {--} \\
	\textbf{\mbox{UnFlow-CS} (ours)}			& 3.30 & {--} & 1.26 & {--} & 8.14 & 23.54\si{\percent} & {--} & 0.65 & {--} & 7.92 & {--} \\
	\textbf{\mbox{UnFlow-CSS} (ours)}			& 3.29 & {--} & 1.26 & {--} & 8.10 & 23.27\si{\percent} & {--} & 0.65 & {--} & 7.91 & 10.22 \\
	\textbf{\mbox{UnFlow-CS-ft} (ours)}		& (1.32) & 1.9 & (0.75) & 0.9 & (2.25) & (9.24\si{\percent}) & 12.55\si{\percent} & 0.64 & {--} & 11.99 & {--} \\
	\textbf{\mbox{UnFlow-CSS-ft} (ours)}	& (1.14) & 1.7 & (0.66) & 0.9 & (1.86) & (7.40\si{\percent}) & 11.11\si{\percent} & 0.64 & 0.76 & 13.65 &{--} \\
  \bottomrule
\end{tabular}
}
\caption{
Accuracy comparison on KITTI, Middlebury, and Sintel optical flow benchmarks.
AEE: Average Endpoint Error; Fl-all: Ratio of pixels where flow estimate is
wrong by both $\geq 3$ pixels and $\geq 5\%$.
The numbers in parentheses are the results of the networks on data they were trained on,
and hence are not directly comparable to other results.
\textsuperscript{\textdagger}\emph{train} numbers are quoted from \citet{Ilg:2017:FEO}, published
before the recent, small update of the KITTI 2015 ground truth. For example, with this update the \emph{test} Fl-all for FlowNet2-ft-kitti changed from 11.48\% to 10.41\%.
}
\label{table:benchmarks}
\end{table*}
}

\subsubsection{Evaluation.}
Images from the KITTI flow benchmarks have a resolution of $1241\times376$, $1226\times370$, or $1242\times375$.
As FlowNet needs both image dimensions to be divisible by $2^6$, we
bilinearly upsample these input images to $1280\times384$ for accuracy evaluation.
The resulting $1280\times384$ flow estimates are then bilinearly downscaled to the original size
for comparison with the ground truth flow maps.
As distances of the original pixels change after resampling, we scale the components of the
estimated flow according to the image downscaling factor in the respective directions.
Note that just zero-padding the input images up to the next valid size introduces visible
artifacts at the image boundaries and significantly increases the error.
For other datasets, we use bilinear resampling analogously.

\begin{figure*}
  \centering
  \includegraphics[width=\textwidth]{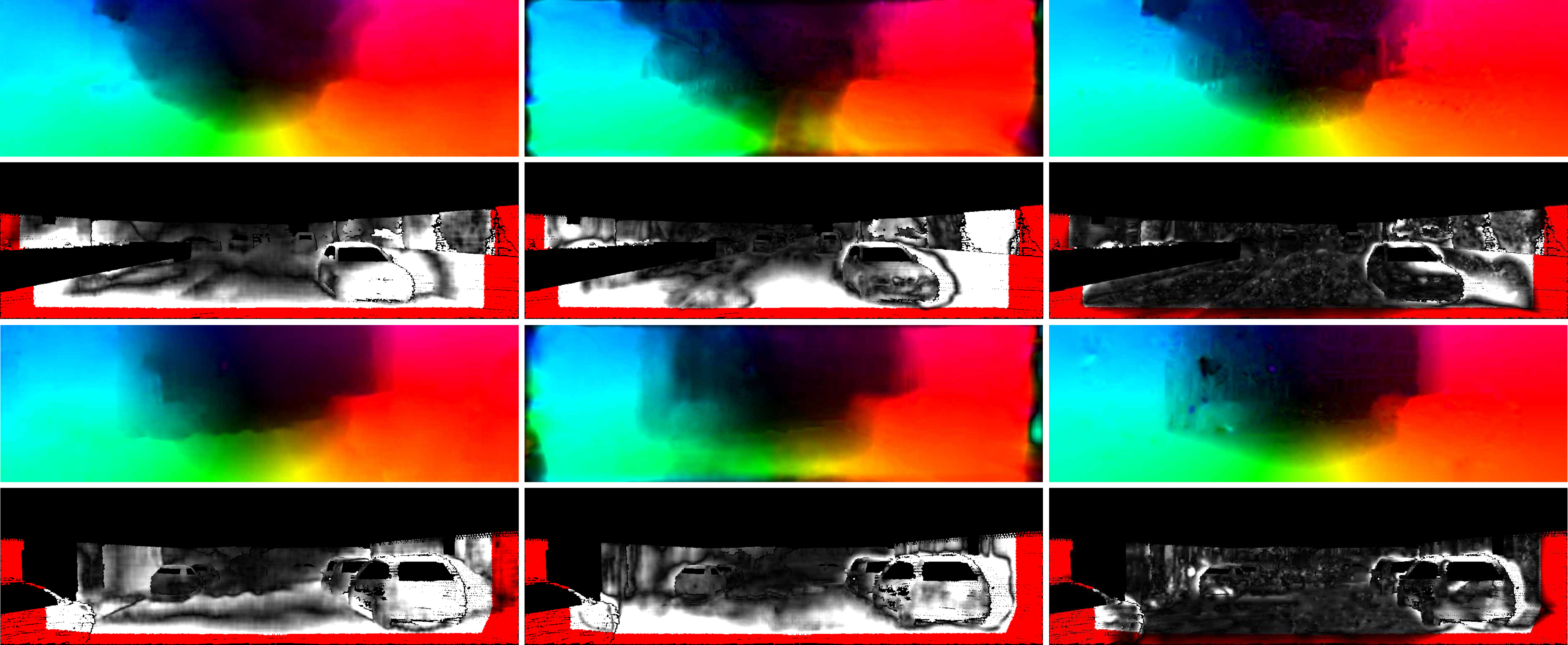}
\caption{
Visual comparison on the KITTI 2012 test set: (left to right) FlowNetS \cite{Dosovitskiy:2015:FLO}, UnsupFlownet \cite{Yu:2016:BBU}, and our unsupervised \mbox{UnFlow-CSS}.
For each example, we show the flow (upper row) and flow error (lower row) maps.
The KITTI 2012 error map scales linearly between 0 (black) and $\geq 5$ pixels error (white).
}
\label{figure:results}
\end{figure*}

\subsection{Unsupervised loss comparison}

To compare the effect of individual unsupervised loss terms, we first pre-train \mbox{UnFlow-C} on SYNTHIA using the census loss, second-order smoothness, and disabling occlusion handling and forward-backward consistency.
We then continue unsupervised training on KITTI for various loss settings.

Table \ref{table:losscomparison} compares the accuracy of different unsupervised losses on the training portions of the KITTI benchmarks.
First, note that when using the exact training protocol of \citet{Yu:2016:BBU},
our re-implementation of their simple brightness constancy baseline loss (with the small change of being formulated bidirectionally)
already achieves higher accuracy than their original implementation (\cf~Table \ref{table:benchmarks}),
which we ascribe to implementation details.

We then observe the effect of each modification of the unsupervised loss terms over our baseline \cite{Yu:2016:BBU} re-implementation.
Our census loss significantly improves upon the original brightness constancy loss ($\sim\!\!\!35\%$ improvement), the second-order smoothness loss clearly outperforms the first-order one ($\sim\!\!5\%$ improvement, $\sim\!\!17\%$ outlier reduction), and occlusion masking combined with forward-backward consistency decreases the overall error further ($\sim\!\!14\%$ improvement).
The combination of all three innovations thus decreases the average endpoint error (AEE, average Euclidean distance between ground truth and estimated flow) by a significant margin compared to previous unsupervised training approaches (to less than $0.5\times$).

It is important to note here that the AEE is not known to the network during training; the only information used in training comes from our unsupervised loss in Eq.~\eqref{eq:1}.

Figure \ref{figure:losscomparison}
visually compares models trained with a baseline loss akin to \cite{Yu:2016:BBU} (\nth{1} and \nth{2} row of Table \ref{table:losscomparison})
and our best unsupervised loss (last row).

\subsection{Results on benchmarks}

We use the best loss setup from the previous ablation study for unsupervised training of our final networks
and refer to our single FlowNetC, two-network stack, and three-network stack trained on SYNTHIA and KITTI images as \mbox{UnFlow-C},
\mbox{UnFlow-CS}, and \mbox{UnFlow-CSS}, respectively.
Finally, we fine-tune \mbox{UnFlow-CS} and \mbox{UnFlow-CSS} with the supervised KITTI schedule and refer to these models as
\mbox{UnFlow-CS-ft} and \mbox{UnFlow-CSS-ft}.
In addition, we train a FlowNetC on SYNTHIA and Cityscapes and refer to this as \mbox{UnFlow-C-Cityscapes}.
Table \ref{table:benchmarks} compares the accuracy of our method
and other optical flow methods on KITTI 2012, KITTI 2015, Middlebury, and MPI Sintel.
Figure \ref{figure:results} visually compares FlowNet-based methods on KITTI 2012.

\subsubsection{KITTI.}
UnsupFlownet \cite{Yu:2016:BBU} and DSTFlow \cite{Ren:2017:UDL} gave some improvements over the supervised FlowNetS \cite{Dosovitskiy:2015:FLO} in non-occluded regions at the cost of a significant increase in total endpoint error on KITTI 2012.
On KITTI 2012, our purely unsupervised \mbox{UnFlow-C} outperforms the \emph{supervised} FlowNetC and FlowNetS in all metrics.
This highlights the benefit of being able to train on the relevant domain, which is possible with our approach even when no ground truth flow is available.
When compared to UnsupFlownet and DSTFlow, \mbox{UnFlow-C} more than halves both error metrics on KITTI 2012
and strongly improves accuracy on KITTI 2015.
Even our unsupervised \mbox{UnFlow-C-Cityscapes}, which was not trained on KITTI images, significantly outperforms UnsupFlownet and DSTFlow.
This shows how our bidirectional loss based on a robust data loss significantly improves over previous unsupervised approaches.
Finally, we observe that our fine-tuned network performs similar to the more complex FlowNet2-ft-kitti \cite{Ilg:2017:FEO} without the need for a separate small-displacement network and custom training schedules.
Note that we do not make use of the extended synthetic dataset of \cite{Mayer:2016:LDT}, but rely on the datasets only as described above.
Moreover, our purely unsupervised networks clearly outperform the supervised FlowNet2 counterparts before fine-tuning.
This again highlights the benefit of unsupervised learning of optical flow for coping with real-world domains.

\subsubsection{Middlebury.}
On Middlebury, \mbox{UnFlow-C} and {UnFlow-C-Cityscapes} outperform the supervised FlowNetS,
and our stacked and fine-tuned variants perform between FlowNetS and the complex FlowNet2 models.
This demonstrates that our networks generalize to realistic domains outside the driving setting they were trained on.

\subsubsection{Sintel.}
On Sintel, our unsupervised networks cannot compete with FlowNetS+ft and FlowNet2, which in contrast are trained on various datasets in a supervised manner.
However, our method outperforms DSTFlow \cite{Ren:2017:UDL}, which is also trained on a similar dataset in an unsupervised manner.
This shows that our method not only strongly outperforms previous unsupervised deep networks for in-domain data, but also yields benefits for data from a domain (here, synthetic) it has not been trained on.

\section{Conclusion}
\label{sec:conclusion}

We presented an end-to-end unsupervised learning approach to enable effective training of FlowNet networks on large datasets for which no optical flow ground truth is available.
To that end, we leveraged components from well-proven energy-based flow approaches, such as a data loss based on the census transform, higher-order smoothness, as well as occlusion reasoning enabled by bidirectional flow estimation.
Our experiments show that using an accurate unsupervised loss, as proposed, is key to exploiting unannotated datasets for optical flow, more than halving the error on the challenging KITTI benchmark compared to previous unsupervised deep learning approaches.
Consequently, we make CNNs for optical flow applicable to a larger range of domains.
Our results, moreover, demonstrate that using a large, real-world dataset together with our unsupervised loss can even outperform supervised training on challenging realistic benchmarks when only handcrafted synthetic datasets are available for supervision.
Finally, we showed that our unsupervised loss provides a solid foundation for pre-training when only limited amounts of real-world ground truth data are available.
Going forward,
our results suggest that further research on more accurate losses for unsupervised deep learning
may be a promising direction for advancing the state-of-the-art in optical flow estimation.

\section{Acknowledgements}
The research leading to these results has received funding from the European Research Council under the European Union's Seventh Framework Programme (FP7/2007--2013) / ERC Grant Agreement No.~307942.

\bibliography{references}
\bibliographystyle{aaai}
\end{document}